# THE KINEMATIC DESIGN OF A 3-DOF HYBRID MANIPULATOR

D. CHABLAT, P. WENGER, J. ANGELES*
*Institut de Recherche en Cybernétique de Nantes (IRCyN)*
*1, Rue de la Noë - BP 92101 - 44321 Nantes Cedex 3 - France*
*Damien.Chablat@ircyn.ec-nantes.fr*
*\* McGill University,*
*817 Sherbrooke Street West, Montreal, Quebec, Canada H3A 2K6*

**Abstract**

This paper focuses on the kinematic properties of a new three-degree-of-freedom hybrid manipulator. This manipulator is obtained by adding in series to a five-bar planar mechanism (similar to the one studied by Bajpai and Roth [1]) a third revolute passing through the line of centers of the two actuated revolute joints of the above linkage (Figures 2 & 3). The resulting architecture is hybrid in that it has both serial and parallel links. Fully-parallel manipulators are known for the existence of particularly undesirable singularities (referred to as parallel singularities) where control is lost [4] and [6]. On the other hand, due to their cantilever type of kinematic arrangement, fully serial manipulators suffer from a lack of stiffness and from relatively large positioning errors. The hybrid manipulator studied is intrinsically stiffer and more accurate. Furthermore, since all actuators are located on the first axis, the inertial effects are considerably reduced. In addition, it is shown that the special kinematic structure of our manipulator has the potential of avoiding parallel singularities by a suitable choice of the « working mode », thus leading to larger workspaces. The influence of the different structural dimensions (e.g. the link lengths) on the kinematic and mechanical properties are analysed in view of the optimal design of such hybrid manipulators.

## 1. Introduction

It is worth noting that most industrial manipulators have a serial kinematic architecture, that is, their links are arranged serially, resulting in a cantilever type structure. The serial kinematic arrangement of links produces large workspaces but suffers from high positioning errors and poor stiffness. Another well know but far less widespread kinematic architecture is the fully-parallel one for which the output link is connected to the ground through several « legs ». Fully-parallel kinematic architectures are known for their high stiffness and low positioning errors but suffer for relatives small workspace. They had been used for a long time in flight simulators and, more recently, in robotic and machine tool applications. The idea of mixing the above two kinds of kinematic arrangement in a single « hybrid » architecture is more recent and still too poorly exploited, despite of the potential benefits which can be gained with such architecture.

This paper focuses on the kinematic properties of a new three-degree-of-freedom hybrid manipulator. This manipulator is obtained by adding in series to a five-bar planar mechanism a third revolute passing through the line of centers of the two actuated revolute joints of the above linkage (Figures 2 & 3). The resulting architecture is hybrid in that it has both serial and parallel links. The hybrid manipulator studied is intrinsically stiffer and more accurate than the classical 3-revolute-jointed 3-DOF fully serial « elbow » manipulator (Figure 1). Furthermore, since all actuators are located on the first axis, the inertial effects are considerably reduced. In addition, it is shown that the special kinematic structure of our manipulator has the potential of avoiding parallel singularities by a suitable choice of the « working mode », thus leading to larger workspaces.

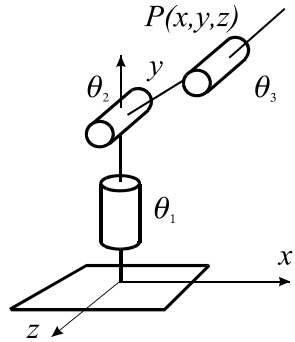

*Figure 1.* The « elbow » manipulator Robot anthropomorphe

The main kinematic equations of the hybrid manipulator are derived. Two Jacobian matrices appear in the kinematic relations between the joint-rate and the Cartesian-velocity vectors, which are called the « inverse-kinematics » and the « direct-kinematics matrices ». These matrices enable the determination of the serial and parallel singularities, which govern the global behavior of the manipulator and define the topological structure of the workspace in Cartesian and joint spaces. The influence of the different structural dimensions (e.g. the link lengths) on the kinematic and mechanical properties are analyzed in view of the optimal design of such hybrid manipulators.

## 2. Preliminaries

### 2.1 A TWO-DOF CLOSED-CHAIN MANIPULATOR

The closed-chain manipulator which constitutes the parallel array of the hybrid manipulator is a five-bar, revolute (*R*)-coupled linkage, as displayed in figure 2. The actuated joint variables are $\theta_1$ and $\theta_2$, while the Cartesian variables are the (*x*, *y*) coordinates of the revolute center *P*. To decrease the inertia effects of moving bodies, we place actuated joints of the closed-chain onto pivots *A* and *B*. Lengths $L_0$, $L_1$, $L_2$, $L_3$, and $L_4$ define the geometry of this manipulator entirely. However, in this paper we focus on a symmetrical manipulators, with $L_1 = L_3$ and $L_2 = L_4$. The symmetrical architecture of the manipulator at hand is justified for general tasks. In manipulator design, then, one is interested in obtaining values of $L_0$, $L_1$, and $L_2$ that optimize a given objective function under some prescribed constraints.

### 2.2 A THREE-DOF HYBRID MANIPULATOR

Now we add one degree of freedom to the manipulator of figure 2. We do this by allowing the overall two-dof manipulator to rotate about line *AB* by means of a revolute coupling the fixed link of the above manipulator with the base of the new manipulator. We thus obtain the manipulator of figure 3.

# THE KINEMATIC DESIGN OF A 3-DOF HYBRID MANIPULATOR

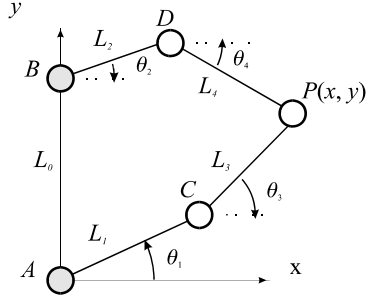
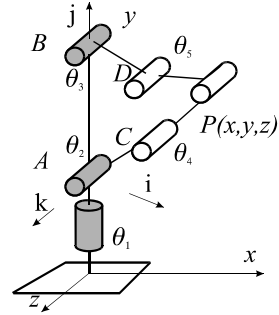

*Figure 2.* A two-dof closed-chain manipulator     *Figure 3.* The three-dof hybrid manipulator

## 2.3 KINEMATIC RELATIONS

The velocity $\dot{\mathbf{p}}$ of point $P$ can be obtained in two different forms, depending on the direction in which the loop is traversed, namely,

$$\dot{\mathbf{p}} = \dot{\mathbf{c}} + (\dot{\theta}_1 \mathbf{j} + \dot{\theta}_4 \mathbf{k}) \times (\mathbf{p} - \mathbf{c}) \quad (1a)$$

and

$$\dot{\mathbf{p}} = \dot{\mathbf{d}} + (\dot{\theta}_1 \mathbf{j} + \dot{\theta}_5 \mathbf{k}) \times (\mathbf{p} - \mathbf{d}) \quad (1b)$$

The equations (1a & b) can be shown to lead to the following 3-dimensional vector equation relating joint and Cartesian velocities [3],

$$\mathbf{A}\dot{\mathbf{p}} = \mathbf{B}\dot{\boldsymbol{\theta}} \quad (2)$$

where

$$\mathbf{A} \equiv \begin{bmatrix} L_2 \mathbf{k}^T \\ (\mathbf{p} - \mathbf{c})^T \\ (\mathbf{p} - \mathbf{d})^T \end{bmatrix} \quad (3)$$

$$\mathbf{B} \equiv L_1 L_2 \begin{bmatrix} \sin(\theta_2) + \lambda_1 \sin(\theta_4) & 0 & 0 \\ 0 & \sin(\theta_2 - \theta_4) & 0 \\ 0 & 0 & \sin(\theta_3 - \theta_5) \end{bmatrix} \quad (4)$$

with $\lambda_1$ defined as $\lambda_1 \equiv L_2 / L_1$, while vectors $\dot{\boldsymbol{\theta}}$ and $\dot{\mathbf{p}}$ are given by

$$\dot{\boldsymbol{\theta}} = [\dot{\theta}_1, \ \dot{\theta}_2, \ \dot{\theta}_3]^T, \ \dot{\mathbf{p}} = [\dot{x}, \ \dot{y}, \ \dot{z}]^T \quad (5)$$

## 2.4 THE WORKING MODE

The manipulator under study has a diagonal inverse-kinematics matrix **B**, as shown in eq. (4), the vanishing of one of its diagonal entries thus indicating the occurrence of a *serial singularity*. The set of manipulator postures free of this kind of singularity is termed a *working mode*. The different working modes are thus separated by a serial singularity, with a set of postures in different working modes corresponding to an inverse kinematic solution. The formal definition of the working mode is detailed in [2]. For the manipulator at hand, the closed chain admits eight working modes, four of which are depicted in figure 4, the remaining four being the mirror image of the first ones about line *AB*.

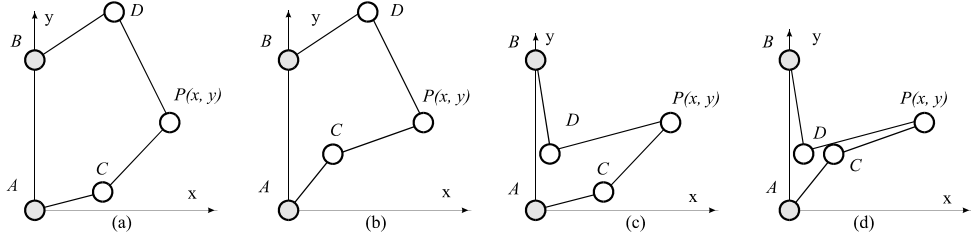

*Figure 4.* The four working modes

## 2.5 THE PARALLEL SINGULARITIES

Parallel singularities occur when the determinant of the direct kinematics matrix **A** vanishes, the corresponding singular configurations occurring inside the workspace. They are particularly undesirable because the manipulator can not resist efforts in some direction(s). To interpret these singularities, we calculate $\mathbf{A}\,\mathbf{A}^T$:

$$\mathbf{A}\,\mathbf{A}^T \equiv L_2^2 \begin{bmatrix} 1 & 0 & 0 \\ 0 & 1 & \cos(\theta_4 - \theta_5) \\ 0 & \cos(\theta_4 - \theta_5) & 1 \end{bmatrix} \quad (6)$$

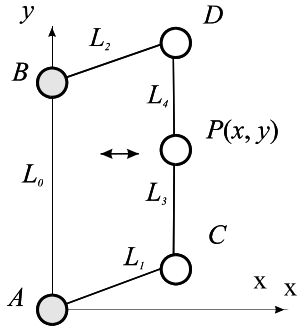 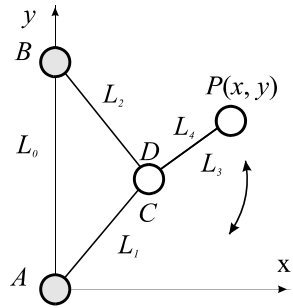

*Figure 5.* Parallel singularity    *Figure 6.* Parallel singularity

whence it is apparent that these singularities occur whenever $\theta_4 - \theta_5 = k\,(\pi/2)$, for an integer $k$, i.e., when the points $C$, $D$ and $P$ are aligned. In such configurations, the manipulator cannot resist an effort in the orthogonal direction of *DC* (Figure 5). Besides, when the points $C$ and $D$ coincide, the position of $P$ is no longer controllable since $P$ can rotate freely around $D$ even if the actuated joints are locked (Figure 6). It will be shown in § 3.2 and § 3.3 that parallel singularities can be avoided either by design or by path-planning.

## 2.6 THE SERIAL SINGULARITIES

Serial singularities occur when the determinant of the inverse kinematics matrix **B** vanishes. When the manipulator is in such a singularity there exists a direction along which no Cartesian velocity can be produced (Figure 7). Such singularities are reached at the boundary of the workspace when *BD* and *DP* or *AC* and *CP* are aligned. They cannot be avoided.

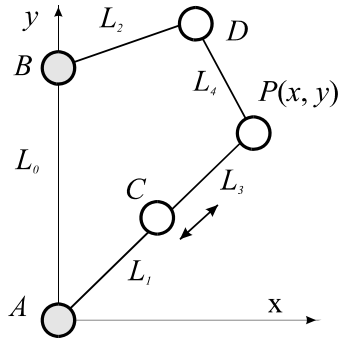

*Figure 7.* Serial singularity

## 3. Kinematic analysis and design

### 3.1 WORKSPACE

A possible design of the manipulator at hand is obtained upon maximizing its workspace volume, which is given by the intersection of two hollow spheres, where radii are $L_1 + L_2$ and $L_1 - L_2$, the distance between the centers of these spheres being $L_0$. Therefore, the workspace is maximum when $L_0$ is equal to $0$ and, by extrapolating the results from [7], when $L_1$ is equal to $L_2$. A cross section of the workspace of such an optimized manipulator is displayed in both Cartesian and joint spaces in figures 8 and 9.

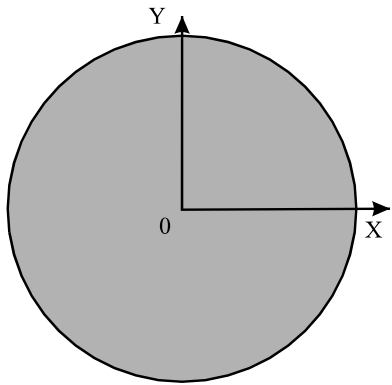
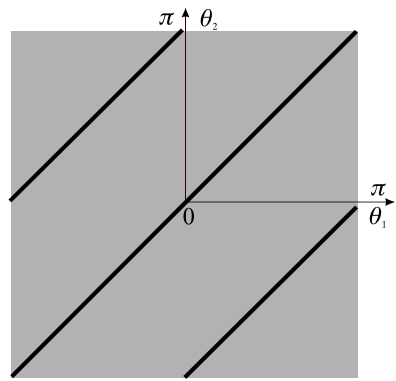

*Figure 8.* A cross section of the Cartesian workspace for $L_0 = 0$     *Figure 9.* A cross section of the joint space

We notice that the optimum manipulator has only two operative working mode because the other two working modes lie in parallel singular configurations (see figures 4 b & c when $L_0 = 0$ and $L_1 = L_2$). Thus, all singularities of this mechanism lie at the border of the workspace.

## 3.2 AVOIDING PARALLEL SINGULARITIES BY PROPER DESIGN

We choose lengths allowing to avoid this configurations for which points $C$ and $D$ coincide. If the sum of lengths $L_1 + L_1$ is greater than the distance between the two pivots $A$ and $B$, i.e. $L_0$, we eliminate a parallel singularity type. In the same way, we can eliminate all parallel singularities if the condition below is imposed:

$$2L_2 - 2L_1 > L_0$$

In using this property, the mechanism has no parallel singularity (Figures 10 & 11).

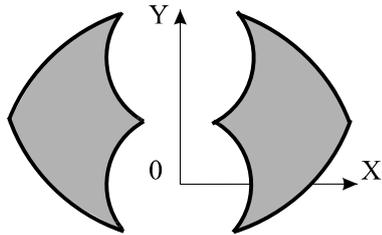 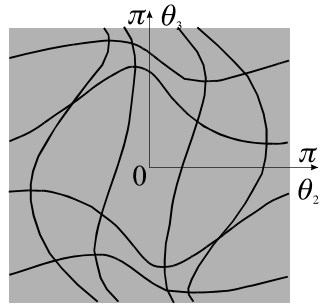

*Figure 10.* Cross-section of the workspace free of parallel singularities

*Figure 11.* Cross-section of the joint space free of parallel singularities

## 3.3 AVOIDING PARALLEL SINGULARITIES BY A CHANGE OF WORKING MODE

According to the working mode chosen, the parallel-singularity posture changes inside the workspace, as displayed in figure 12.

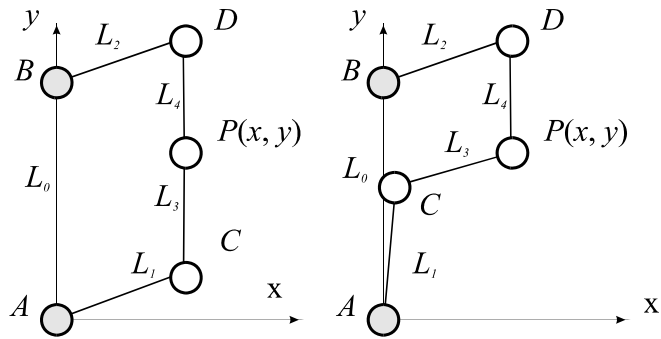

*Figure 12.* Working mode and parallel singularity

In the « convex working mode » (Figure 12 left), the manipulator is in parallel singularity, while it is in a regular configuration in the non-convex working mode (Figure 12 right). Also, to execute operations in a highly constrained environment, it can be desirable to change working mode. It is possible, by means of a change of working mode, to reach points inside the two workspace sections displayed in figures 13 and 14. To change working mode, it is necessary to pass through a serial singularity, located at the boundary of the Cartesian workspace. By doing so, we can reach a workspace that is twice as large as each of those displayed in figures 13 and 14. In fact, parallel singularities not being avoidable, the workspace of a parallel manipulator is always smaller than that of its serial counterpart of the same size. Also, according to the working mode, the condition numbers of the direct- and the inverse-kinematics matrices are different. Thus, according to the task to execute, requiring either a precise motion or a fast motion, we can choose one suitable working mode.

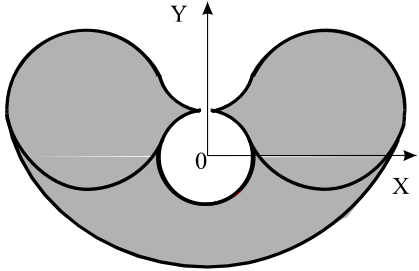 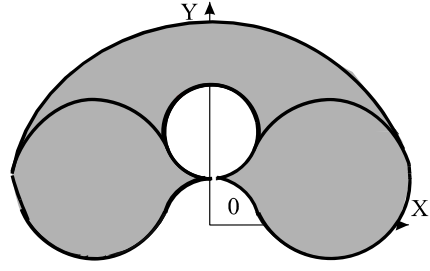

*Figure 13.* Workspace of the convex working mode

*Figure 14.* Workspace of the nonconvex working mode

## *3.4 THE ISOCONDITIONING CURVES*

Another interesting design criteria is the isoconditioning curves in the workspace ([5] and [3]).
We derive below the loci of equal condition number of the direct-kinematics matrices. To do this, we first recall the definition of *condition number* of an $m \times n$ matrix **M**, with $m \leq n$, $\kappa(\mathbf{M})$. This number can be defined in various ways; for our purposes, we define $\kappa(\mathbf{M})$ as the ratio of the largest, $\sigma_l$, to the smallest $\sigma_s$, singular values of **M**, namely,

$$\kappa(\mathbf{M}) = \frac{\sigma_l}{\sigma_s} \quad (7)$$

The singular values $\{\sigma_k\}_1^m$ of matrix **M** are defined, in turn, as the square roots of the nonnegative eigenvalues of the positive-semidefinite $m \times m$ matrix $\mathbf{M}\mathbf{M}^T$.
The eigenvalues of the matrix $\mathbf{A}\mathbf{A}^T$ (6) are, $\alpha_1 = 1 - |\cos(\theta_4 - \theta_5)|$, $\alpha_2 = 1$, and $\alpha_3 = 1 + |\cos(\theta_4 - \theta_5)|$ [3], the foregoing eigenvalues having been ordered as

$$\alpha_1 \leq \alpha_2 \leq \alpha_3$$

The condition number of matrix **A** is thus

$$\kappa(\mathbf{A}) = \frac{1}{|\tan((\theta_4 - \theta_5)/2)|} \quad (8)$$

These loci are, in fact, surfaces of revolution generated by the isoconditioning curves of the 2-dof manipulator, when these are rotated about the axis of the first revolute.
In light of expression (8) for the condition number of the Jacobian matrix **A**, it is apparent that $\kappa(\mathbf{A})$ attains its minimum of 1 when $|\theta_3 - \theta_4| = \pi/2$, the equality being understood *modulo* $\pi$. At the other end of the spectrum, $\kappa(\mathbf{A})$ tends to infinity when $\theta_3 - \theta_4 = k\pi$, for *k= 1, 2, ...* . When matrix **A** attains a condition number of unity, it is termed *isotropic*, its inversion being performed without any roundoff-error amplification. Manipulator postures for which condition $\theta_3 - \theta_4 = \pi/2$ holds are thus the most accurate for purposes of the direct kinematics of the manipulator. Correspondingly, the locus of points whereby matrix **A** is isotropic is called the *isotropy locus* in the Cartesian workspace.
On the other hand, manipulator postures whereby $\theta_3 - \theta_4 = k\pi$ denote a singular matrix **A**, and hence, define the boundary of the Cartesian workspace of the manipulator. Such singularities occur at the boundary of the Cartesian workspace of the manipulator, and hence, the locus of *P* whereby these singularities occur, namely, the *singularity locus* in the Cartesian space, defines this boundary.

Interestingly, isotropy can be obtained regardless of the dimensions of the manipulator, as long as *i)* it is symmetric and *ii)* $L_2 \neq 0$. For the two working modes, the position of the isotropic configurations are different (Figure 15).

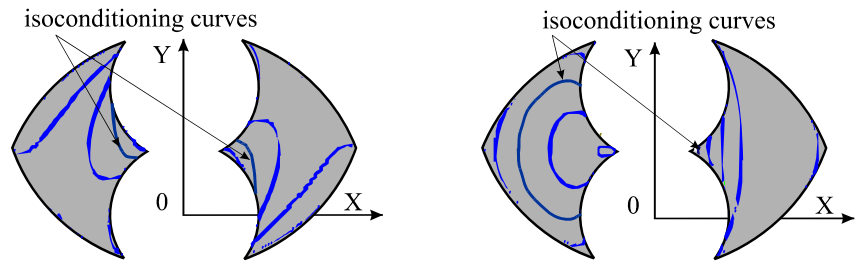

*Figure 15.* Isoconditioning curves for two working modes

The isoconditioning curves are, by definition, the sets of points which are at equal distance from the singularities and, in turn, can be used as a criterion for designing and/or placing the manipulator in such a way that the working trajectories be in a region of the workspace where a minimum stiffness is guaranteed.

## 4. Conclusions

We have defined a novel architecture of hybrid manipulators and produced design and path-planning guidelines for the implementation of these manipulators. Two Jacobian matrices were identified in the mapping of joint rates into Cartesian velocities, namely, the direct-kinematics and the inverse-kinematics matrices.

The hybrid manipulators studied were optimized for maximum Cartesian workspace volume and iso-conditioning. Further research work is being conducted by the authors on such hybrid manipulators with regard to their optimal design.

## 5. Acknowledgments

The third author acknowledges the support from the Natural Sciences and Engineering Research Council, of Canada, the Fonds pour la formation de chercheurs et l'aide à la recherche, of Quebec, and École Centrale de Nantes (ECN). The research reported here was conducted during a sojourn that this author spent at ECN's Institut de Recherche en Cybernétique de Nantes.

**References**

[1] Bajpai A., Roth B. : Workspace and mobility of a closed-loop manipulator, *The International Journal of Robotics Research*, Vol. 5, No. 2, 1986.

[2] Chablat D., Wenger Ph. : Domaines d'unicité des manipulateurs parallèles : Cas général, *IRCyN Internal Report*, No. 97.7, École Centrale de Nantes, Nantes, 1997.

[3] Chablat D., Wenger Ph., Angeles J. : The isoconditioning Loci of A Class of Closed-Chain Manipulators, *Proc. IEEE International Conference of Robotic and Automation*, pp. 1970-1976, Mai 1998.

[4] Gosselin C., Angeles J. : Singularity analysis of closed-loop kinematic chains, *Proc. IEEE Transactions On Robotics And Automation*, Vol. 6, No. 3, June 1990.

[5] Gosselin C. : Stiffness Mapping for Parallel Manipulators, *Proc. IEEE Transactions On Robotics And Automation*, Vol. 6, No. 3, June 1990.

[6] Merlet J-P. : Les robots parallèles, HERMES, seconde édition, Paris, 1997.

[7]   Paden B., Sastry S. : Optimum kinematic design of 6R manipulators, *The International Journal of Robotic Research*, Vol. 7, 1988, No. 2, pp. 43-61.